\pdfoutput=1

\documentclass[11pt]{article}
\usepackage{amsmath,amssymb}  
\usepackage[final]{acl}
\usepackage{adjustbox}
\usepackage{times}
\usepackage{latexsym}
\usepackage{booktabs}
\usepackage{siunitx}
\usepackage{enumitem} 

\usepackage{tikz}
\usetikzlibrary{mindmap,trees}

\usepackage[T1]{fontenc}

\usepackage[utf8]{inputenc}
\usepackage{graphicx}

\usepackage{microtype}
\usepackage{svg}
\usepackage{inconsolata}

\usepackage{graphicx}

\title{PediaMind-R1: A Temperament-Aware Language Model for Personalized Early Childhood Care
 Reasoning via Cognitive Modeling and Preference Alignment}

\author{
Zihe Zhang\textsuperscript{1},
Can Zhang\textsuperscript{1},
Yanheng Xu\textsuperscript{2},
Xin Hu\textsuperscript{1},
Jichao Leng\textsuperscript{1} \\
\textsuperscript{1}School of Future Information and Innovation, Fudan University, Shanghai, China \\
\textsuperscript{2}Corporate Research, Bosch (China) Investment Ltd., Shanghai, China \\
}

\begin{document}
\maketitle
\begin{abstract}
This paper presents PediaMind-R1, a domain-specialized large language model designed to achieve active personalization in intelligent parenting scenarios. Unlike conventional systems that provide generic suggestions, PediaMind-R1 draws on insights from developmental psychology. It introduces temperament theory from the Thomas–Chess framework and builds a temperament knowledge graph for infants and toddlers (0–3 years). Our two-stage training pipeline first uses supervised fine-tuning to teach structured chain-of-thought reasoning, and then applies a GRPO-based alignment stage to reinforce logical consistency, domain expertise, and empathetic caregiving strategies. We further design an evaluation framework comprising temperament-sensitive multiple-choice tests and human assessments. The results demonstrate that PediaMind-R1 can accurately interpret early childhood temperament profiles and proactively engage in individualized reasoning. This work highlights the value of integrating vertical-domain modeling with psychological theory. It offers a novel approach to developing user-centered LLMs that advance the practice of active personalization in sensitive caregiving contexts.
\end{abstract}

\section{Introduction}

Large language models (LLMs) have shown strong general performance across diverse tasks. However, most are designed for generic usage and lack the ability to adapt to individual users. Both active and passive personalization, whether guided by user input or inferred from interaction history, remain underdeveloped, with limited ability to condition responses on structured user characteristics.

Personalization is especially critical in domains such as parenting and infant care, where individual needs vary widely and generic suggestions may be insufficient. Developmental psychology has long emphasized that caregiving strategies tailored to a child's temperament, including traits such as adaptability and emotional intensity, can significantly impact long-term developmental outcomes. The Thomas--Chess temperament model, for example, categorizes infants into structured types such as ``easy,'' ``difficult,'' and ``slow-to-warm-up,'' offering a psychologically grounded basis for individualization \citep{thomas1977temperament, carey2004temperament}.

In this work, we propose PediaMind-R1, a domain-specialized LLM for active personalization in early childhood care. Building on the Thomas--Chess framework, we construct a temperament knowledge graph and condition model outputs on temperament labels to deliver individualized caregiving strategies. Our two-stage training pipeline combines supervised fine-tuning (SFT) \citep{hu2022lora} for structured reasoning with Group Relative Policy Optimization (GRPO) \citep{zhang2025r1vl} which selects responses exceeding group-average performance. To evaluate effectiveness, we design a temperament-sensitive framework with scenario-based multiple-choice tests and expert assessments, confirming the model’s ability to interpret temperament and provide personalized, empathetic recommendations.

Our contributions are threefold:
\begin{itemize}
  \setlength\itemsep{0pt}
  \setlength\parsep{0pt}
  \setlength\parskip{0pt}

    \item \textbf{Activating LLM Personalization via Psychological Temperament Modeling:} 
We leverage temperament traits from the Thomas--Chess framework to explicitly model psychological profiles, thereby activating personalized reasoning in early childhood care and aligning LLM outputs with children’s unique developmental needs.

    \item \textbf{Modeling Temperament-Aware Reasoning via SFT and GRPO:}
    We jointly apply SFT and GRPO to embed temperament-sensitive reasoning into the LLM, combining structured logic with preference alignment grounded in developmental psychology.

    \item \textbf{Temperament-Sensitive Evaluation:}
    We propose an evaluation scheme using multiple-choice benchmarks and expert assessments to capture both factual accuracy and psychological appropriateness.
\end{itemize}

Although newer temperament frameworks exist, we adopt the classical Thomas--Chess model \citep{thomas1977temperament} as a widely recognized baseline to validate our methodology.

\section{Related Work \& Motivation}

\subsection{Infant Temperament as a Personalization Signal}

Most personalization strategies in artificial intelligence assume that users can explicitly articulate their needs. However, in domains such as infant care, the end user—the infant—lacks communicative agency, necessitating proxy-driven personalization. Among psychological frameworks, the temperament theory proposed by Thomas and Chess \citep{thomas1977temperament} is particularly influential. It categorizes infants based on observable traits such as adaptability, activity level, and emotional intensity.

These temperament classifications have demonstrated predictive value for long-term developmental outcomes and are widely used to inform parenting decisions \citep{carey2004temperament}. Instruments like the Infant Temperament Questionnaire (ITQ) offer a structured way for caregivers to assess these traits. In this work, we use these traits as personalization signals, conditioning LLM reasoning on temperament profiles to support individualized parenting strategies.

\subsection{Personalizing LLMs via Human-Guided Reward Alignment}

Prior efforts in LLM personalization include user embedding approaches \citep{madotto2019personalizing}, in-context learning paradigms \citep{khorashadizadeh2023incontext}, and retrieval-augmented methods \citep{liu2020personalization}. However, these approaches typically rely on explicit user feedback or long-term interaction histories, which are unavailable in non-verbal, high-stakes domains such as infant care.

To address this, we adopt Group Relative Policy Optimization (GRPO), a reinforcement learning method that compares multiple candidate outputs for the same prompt and computes a group-relative advantage. Unlike Direct Preference Optimization (DPO) \citep{rafailov2023dpo}, which depends on binary preference pairs, GRPO evaluates outputs by their relative performance within a group, enabling stable optimization without requiring absolute reward labels \citep{zhang2025r1vl}. 

This strategy is well-suited for temperament-sensitive reasoning, where correctness is graded across dimensions such as logical consistency, psychological alignment, safety, and empathy. Our approach thus combines psychological profiling, curated supervision, and structured reward design to realize active personalization in infant care, drawing inspiration from reasoning-focused models such as DeepSeek-R1 \citep{guo2025deepseekr1, liu2024deepseekv3}.

\section{Methodology}

\begin{figure}[htbp]
\centering
\includegraphics[width=\linewidth]{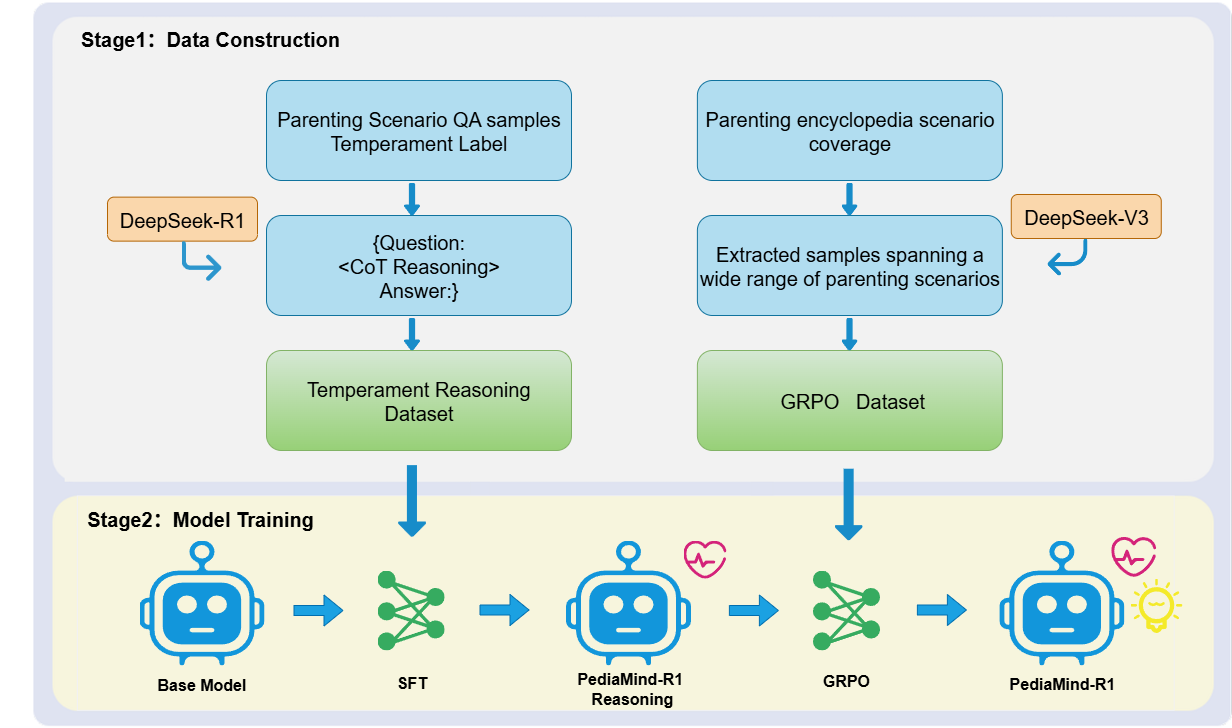}
\caption{PediaMind-R1 training pipeline: temperament-aware supervised fine-tuning, followed by GRPO alignment.}
\label{fig:pipeline}
\end{figure}

We adopt a streamlined two-stage training framework to develop a temperament-sensitive LLM for infant care, as shown in Figure~\ref{fig:pipeline}. Our approach consists of (1) temperament-aware supervised fine-tuning using LoRA and (2) group-relative preference optimization, with detailed training configurations provided in Appendix~B.

\subsection{Temperament-Aware Supervised Fine-Tuning}

\subsubsection{Dataset Construction}
\begin{figure}[h]
    \centering
    \includegraphics[width=0.48\textwidth]{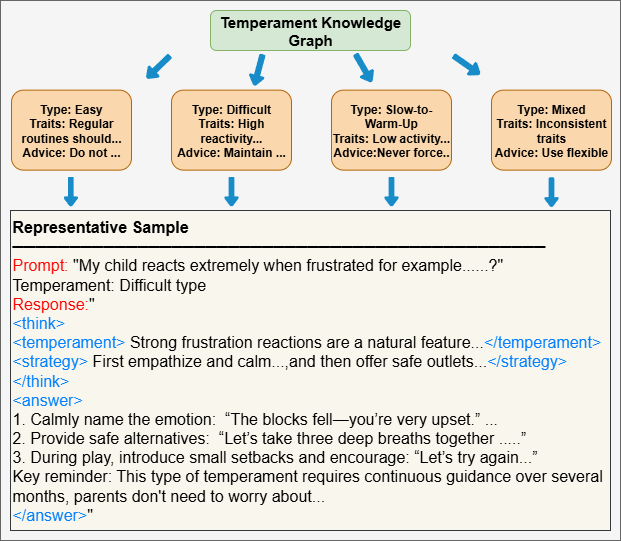}
    \caption{Representative dataset sample combining temperament knowledge graph and structured response.}
    \label{fig:dataset-sample}
\end{figure}

Our supervised fine-tuning dataset comprises 1,215 caregiver queries annotated with temperament labels from the Thomas–Chess framework and structured chain-of-thought responses. Responses were generated with DeepSeek-R1 and guided by a curated temperament–strategy knowledge graph (see AppendixC), with 10\% expert-reviewed for factual and psychological validity. A representative example is shown in Figure\ref{fig:dataset-sample}.

\subsubsection{Supervised Fine-Tuning}
Using this dataset, we fine-tune the base model with LoRA adaptation, enabling it to ground multi-step reasoning and recommendations in explicit temperament profiles. The model produces structured, explainable responses aligned with psychological theory and practical caregiving, enhancing personalization while ensuring transparent chain-of-thought reasoning.

\subsection{Group Relative Preference Optimization}

\subsubsection{GRPO Algorithm}
During this phase, we adopt the GRPO algorithm to update the model based on group-relative advantage. For each scenario, $G$ candidate responses are sampled from the old policy $\pi_{\text{old}}$, each assigned a reward $r_i$. The group-relative advantage for the $i$-th output is calculated as:
{\small
\begin{equation}
A_i = \frac{r_i - \mu_{\{r\}}}{\sigma_{\{r\}}}
\end{equation}
}

where $r_i$ is the reward of the $i$-th output, and $\mu_{\{r\}}$, $\sigma_{\{r\}}$ are the mean and standard deviation of reward values in the group. The policy is optimized to maximize:

{\scriptsize
\begin{align}
\mathcal{J}_{\text{GRPO}}(\theta) =\; & \mathbb{E}_{S} \Bigg[ \frac{1}{G} \sum_{i=1}^{G} \min \Big( r_i A_i,\, \mathrm{clip}(r_i, 1-\epsilon, 1+\epsilon) A_i \Big) \notag \\
& \qquad -\, \beta D_{\mathrm{KL}}(\pi_\theta \Vert \pi_{\text{ref}}) \Bigg]
\end{align}
}

where $D_{\mathrm{KL}}$ is the KL-divergence penalty for regularization. This group-based mechanism encourages the model to generate outputs whose rewards exceed the group average, leading to more robust and stable preference alignment.

\subsubsection{Reward Design}
Each candidate output $y$ is evaluated by a composite reward function, capturing three critical aspects:
\[
\mathcal{R}(y) = R_{\text{fmt}}(y) + R_{\text{temp}}(y) + R_{\text{know}}(y)
\]
where
{\scriptsize
\begin{align}
R_{\text{fmt}}(y) &=
\begin{cases}
1 & \text{if output strictly matches constructed format} \\
0 & \text{otherwise}
\end{cases} \\
R_{\text{temp}}(y) &=
\begin{cases}
1 & \text{if reasoning aligns with temperament knowledge} \\
0 & \text{otherwise}
\end{cases} \\
R_{\text{know}}(y) &=
\begin{cases}
1 & \text{if answer is fully relevant to query/reference} \\
0.5 & \text{if answer is partially relevant to query/reference} \\
0 & \text{otherwise}
\end{cases}
\end{align}
}
This multi-dimensional reward structure promotes outputs that are standardized, temperament-logical, and professionally grounded. GRPO ensures psychologically sound and domain-aligned recommendations. To support this, we built a dataset of 2,646 temperament-sensitive scenarios from a DeepSeek-V3–assisted parenting encyclopedia and a temperament knowledge graph, with 15\% reviewed by pediatric and psychology experts for reliability.

\section{Experiments}

\subsection{Benchmark Evaluation}
We evaluated PediaMind-R1 on 200 temperament-sensitive multiple-choice questions combining infant profiles, caregiving challenges, and candidate strategies. Comparisons were made against the untuned Qwen2.5-7B-Instruct baseline, with ablations for (i) supervised fine-tuning (SFT) and (ii) SFT + GRPO alignment to assess incremental training benefits.

Each model was prompted in a zero-shot multiple-choice format and required to select the single best answer per scenario. An illustrative example is provided below:

\begin{quote}
\textbf{Scenario:}  
When guests visit the home, my child immediately hides in their room and refuses to come out for a long time. Temperament: \textit{slow-to-warm-up}

\textbf{Question:}  
Which of the following parenting strategies is most appropriate for this scenario?

\begin{enumerate}[label=(\Alph*), leftmargin=0em, itemsep=0.8ex]
    \item Wait for the child to adjust and gently invite them to join when comfortable. \textit{(Best fit: slow-to-warm-up temperament)}
    \item Insist that the child come out right away to face social situations directly.
    \item Leave the child alone in their room until they decide to come out.
\end{enumerate}
\end{quote}

\begin{table}[htbp]
\centering
\renewcommand{\arraystretch}{1.3} 
\setlength{\tabcolsep}{10pt} 
\scriptsize 
\begin{tabular}{lcc}
\toprule
\textbf{Model} & \textbf{Size} & \textbf{Accuracy (\%)} \\
\midrule
Qwen2.5-7B-Instruct (untuned) & 7B & 55.0 \\
PediaMind-R1 (SFT only)       & 7B & 62.0 \\
PediaMind-R1 (SFT+GRPO)       & 7B & \textbf{67.0} \\
\bottomrule
\end{tabular}
\caption{Accuracy on temperament-sensitive multiple-choice benchmark (top-1 selection rate) across the Qwen2.5-7B-Instruct baseline and PediaMind-R1 ablation settings.}
\label{tab:ablation}
\end{table}

As shown in Table~\ref{tab:ablation}, temperament-aware supervised fine-tuning (SFT) markedly enhanced PediaMind-R1, confirming its role in instilling structured reasoning and embedding fundamental temperament knowledge. However, we still noted occasional mismatches between behavioral cues and recommended strategies, pointing to the limited breadth of training scenarios.

Subsequent GRPO alignment further reinforced logical consistency and psychological appropriateness, rewarding outputs more closely aligned with developmental psychology principles. Although the absolute gain was modest, GRPO consistently improved logical consistency and psychological appropriateness across diverse scenarios. Overall, these findings suggest that while SFT provides a solid foundation for temperament-sensitive reasoning, GRPO is essential for consolidating robustness and ensuring more reliable personalization.

\subsection{Human Assessment}
We designed 100 scenario-based queries covering diverse infant temperament types and caregiving situations. This evaluation complements the accuracy benchmark by capturing qualitative aspects beyond correctness, such as psychological alignment and caregiving suitability. Three domain experts (developmental psychology PhD, pediatric nursing MSc, and artificial intelligence MSc) conducted a blinded evaluation, where anonymized outputs from different PediaMind-R1 variants were presented side-by-side in randomized order. Each answer was independently rated on (a) knowledge correctness, (b) psychological appropriateness, and (c) caregiving suitability, using a 0--1 scale. Final scores were computed by averaging the three expert ratings for each dimension. Inter-rater agreement among the three experts reached 0.81 (Cohen’s $\kappa$), indicating substantial consistency and reliability of the evaluation process.

\begin{table}[htbp]
\centering
\tiny 
\setlength{\tabcolsep}{6pt} 
\renewcommand{\arraystretch}{1.05} 
\begin{tabular}{lccc}
\toprule
\textbf{Model} & \textbf{Knowledge} & \textbf{Psych. Align.} & \textbf{Caregiving} \\
\midrule
Qwen2.5-7B-Instruct (untuned) & 0.68 & 0.68 & 0.75 \\
PediaMind-R1 (SFT only)       & 0.66 & 0.88 & 0.83 \\
PediaMind-R1 (SFT+GRPO)       & \textbf{0.72} & \textbf{0.92} & \textbf{0.88} \\
\bottomrule
\end{tabular}
\caption{Expert evaluation results on a 0--1 scale across 100 scenario-based queries, covering knowledge correctness, psychological alignment, and caregiving suitability.}
\label{tab:expert_ablation}
\end{table}

As shown in Table~\ref{tab:expert_ablation}, SFT enhanced the model’s temperament-awareness but occasionally produced answers with weak query relevance, suggesting an overemphasis on structured formats. GRPO addressed this limitation by rewarding content fidelity, leading to significant improvements in psychological alignment and caregiving suitability.

\section{Conclusion}

Our study shows that combining psychological profiling with reward-guided training enables effective personalization in infant care. PediaMind-R1 gained temperament reasoning via SFT, while GRPO refined robustness and psychological alignment. This two-stage pipeline offers a reliable framework for extending temperament-sensitive personalization to domains such as healthcare and education. Beyond infancy, the approach highlights the potential of integrating cognitive modeling with reinforcement-based alignment to support other sensitive, user-centered applications requiring both accuracy and empathy.

\section*{Limitations}
Our approach relies on caregiver-provided temperament assessments, which may vary in accuracy due to potential reporting bias. While this study focuses on temperament reasoning accuracy, the supervised dataset remains relatively small, and we employ only the classical Thomas–Chess framework, which may limit coverage of modern temperament models. Moreover, our reward design uses largely discrete signals, and no formal significance testing was conducted due to the modest benchmark size, leaving both finer-grained optimization and broader validation for future work.

\section*{Acknowledgments}
This research was supported by Fudan University and Bosch China, whose funding and expert guidance were essential to this work.
\clearpage
\bibliography{custom}


\clearpage
\appendix
\section{Background on the Thomas--Chess Temperament Framework}
\label{sec:appendixA}

The Thomas–Chess framework \citep{thomas1977temperament} is a seminal model in developmental psychology based on the New York Longitudinal Study (NYLS), initiated in the 1950s, which tracked infants over time to identify stable temperament profiles. From nine behavioral dimensions—including activity level, adaptability, mood quality, and attention persistence—they distilled three predominant temperament categories: Easy, Difficult, and Slow-to-Warm-Up:

\begin{itemize}[leftmargin=1em,itemsep=0.2em]
  \item \textbf{Easy}: Regular biological rhythms, adaptable responses to environmental changes, and predominantly positive affective tone.
  \item \textbf{Difficult}: Irregular biological patterns, low adaptability, high withdrawal or negative emotionality.
  \item \textbf{Slow-to-Warm-Up}: Low activity levels, initial withdrawal when confronted with novelty, but gradual habituation with repeated exposure.
\end{itemize}

Importantly, approximately 35\% of infants did not fit neatly into these three clusters and were categorized as exhibiting a Mixed temperament — a combination of traits crossing multiple dimensions, without dominance of any single profile.

Subsequent work by Carey \citep{carey2004temperament} enhanced the clinical and pediatric application of this model, emphasizing its utility for individualized parenting strategies. The Thomas–Chess typology has also demonstrated predictive validity for socio-emotional and behavioral outcomes later in childhood.

From a computational perspective, the Thomas–Chess framework provides a structured taxonomy of temperament-relevant traits that can be directly operationalized as features or labels in machine learning systems. In PediaMind-R1, we encode temperament profiles—including mixed-type assessments—as conditioning signals, thereby enabling the model to deliver tailored, psychologically grounded recommendations that maintain interpretability and theoretical rigor.

\section{Details of Training Setup}

We provide detailed training configurations for both the Supervised Fine-Tuning (SFT) and Reinforcement Learning (RL) phases of PediaMind-R1. During the SFT stage, we fine-tuned the base model (Qwen2.5-7B-Instruct) using LoRA adaptation. In the RL stage, we employed Group Relative Policy Optimization (GRPO) with a rollout group size of 4, enabling the model to compare multiple candidate outputs for each prompt and optimize relative to the group average. 

All experiments were conducted on an 8$\times$80GB NVIDIA A100 GPU platform. Key hyperparameters for both SFT and GRPO stages are summarized in Table~\ref{tab:training_params}.

\begin{table}[htbp]
\centering
\resizebox{0.8\linewidth}{!}{%
\begin{tabular}{lcc}
\toprule
\textbf{Parameter} & \textbf{SFT} & \textbf{RL(GRPO)} \\
\midrule
Batch Size (per device) & 2 & 4 \\
Gradient Accumulation & 2 & 8 \\
Global Batch Size & 32 & 256 \\
Epochs & 5 & 3 \\
Learning Rate & 2.0e-5 & 1.0e-6 \\
Warmup Ratio & -- & 0.03 \\
Max Prompt Length & -- & 512 \\
Max Completion Length & -- & 1024 \\
Max Sequence Length & 1024 & 1024 \\
Optimizer & AdamW & AdamW \\
Adam $\beta_1$ / $\beta_2$ & -- & 0.9 / 0.99 \\
Weight Decay & 0.01 & 0.1 \\
LR Scheduler & Cosine & Cosine \\
Gradient Checkpointing & True & -- \\
Evaluation / Logging Steps & 10 & 10 \\
Save Steps & 100 & 20 \\
Save Total Limit & 3 & -- \\
Max Grad Norm & -- & 0.5 \\
Temperature & -- & 1.0 \\
Rollout Generations & -- & 4 \\
KL Coefficient ($\beta$) & -- & 0.005 \\
Precision & bfloat16 & bfloat16 \\
\bottomrule
\end{tabular}}
\caption{Training hyperparameters for SFT and GRPO stages, executed on an 8$\times$80GB NVIDIA A100 GPU platform.}
\label{tab:training_params}
\end{table}

\section{Temperament Knowledge Graph}
\label{sec:appendixC}

To provide structured support for temperament-aware reasoning, we summarize the Thomas--Chess temperament framework as a knowledge graph (see Table~\ref{tab:temperament-graph}). The design follows the principle of Goodness of Fit: parenting success depends not on changing the child’s temperament, but on adapting caregiving practices and environments to match it.

\begin{table*}[htbp]
\centering
\renewcommand{\arraystretch}{1.35}
\setlength{\tabcolsep}{5pt}
\scriptsize
\begin{tabular}{p{0.21\linewidth} p{0.21\linewidth} p{0.21\linewidth} p{0.21\linewidth}}
\toprule
\textbf{Easy Child ($\approx 40\%$)} & \textbf{Difficult Child ($\approx 10\%$)} & \textbf{Slow-to-Warm-Up ($\approx 15\%$)} & \textbf{Mixed Type ($\approx 35\%$)} \\
\midrule
\textbf{Traits:} Regular biological rhythms, quick adaptability, moderate reaction intensity, generally cheerful. Children usually adjust smoothly to changes in routines or environments.  
& \textbf{Traits:} Irregular biological rhythms, low adaptability, high withdrawal, intense reactions, more frequent negative mood. These children are easily upset by unfamiliar events or transitions.  
& \textbf{Traits:} Low activity level, initial avoidance of novelty, slow adaptability, mild emotional expressions. They often observe cautiously before joining new activities.  
& \textbf{Traits:} Exhibit a mixture of the other three types, with context-dependent reactions. No single temperament trait dominates, making responses less predictable. \\
\textbf{Advice:}  
1. Do not neglect needs due to ``easy'' behavior.  
2. Attend to subtle emotional signals.  
\textit{Example: Even if the child plays quietly, schedule regular check-ins for comfort and engagement.}  
& \textbf{Advice:}  
1. Maintain calm, consistent parental emotions.  
2. Create a predictable, structured daily routine.  
\textit{Example: Use a visual daily schedule chart to help the child know what comes next and reduce anxiety.}  
& \textbf{Advice:}  
1. Avoid forcing or rushing into new situations.  
2. Act as a ``secure base'' by modeling positive interaction.  
3. Reframe traits positively (e.g., ``cautious'' instead of ``shy'').  
\textit{Example: Introduce new environments gradually—first observing with a parent, then gently participating.}  
& \textbf{Advice:}  
1. Apply a situational ``deconstruction'' approach: assess traits per context.  
2. Flexibly switch strategies to achieve dynamic fit.  
\textit{Example: If the child shows Easy-type reactions at home but Slow-to-Warm-Up at school, adjust parenting accordingly.} \\
\bottomrule
\end{tabular}
\caption{Knowledge graph of infant temperament categories and caregiving strategies, based on the Thomas--Chess framework. Each type is illustrated with traits, tailored advice, and practical examples, all emphasizing the principle of Goodness of Fit.}
\label{tab:temperament-graph}
\end{table*}

\section{Example of PediaMind-R1 Response}
\label{sec:appendixD}

\begin{figure*}[htbp]
    \centering
    \includegraphics[width=0.9\textwidth]{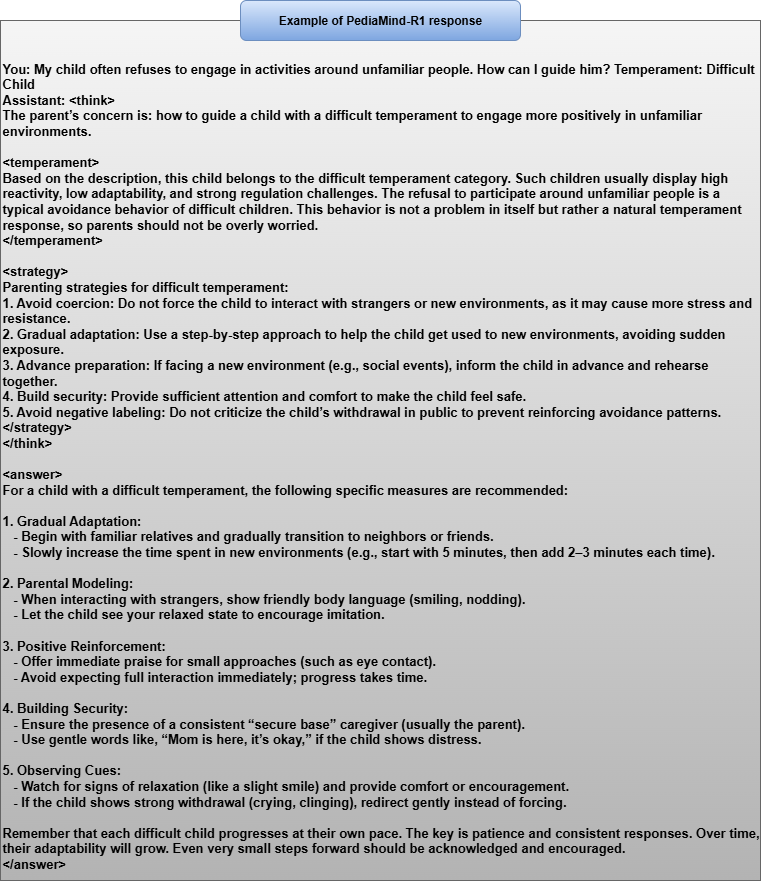}
    \caption{PediaMind-R1 response example for a child with a difficult temperament.}
    \label{fig:example}
\end{figure*}

To illustrate the practical use of our system, Figure~\ref{fig:example} presents an example query and the corresponding PediaMind-R1(finetuned with SFT and GRPO) response. The example demonstrates how temperament profiling (here: \textit{Difficult Child}) guides the reasoning process.

\end{document}